\documentclass{svproc}
\usepackage{url}

\usepackage{graphicx}
\newcommand{\slas}[1]{\mbox{${{#1} \!\!\! /}$}}

\begin{document}
\mainmatter              
\title{A neural Markovian multiresolution image labeling algorithm}
\titlerunning{MCV algorithm}  
%
\author{John Mashford\inst{1} \and Brad Lane\inst{2}
Vic Ciesielski\inst{3} \and Felix Lipkin\inst{2}}
\authorrunning{John Mashford et al.} 
%
\tocauthor{John Mashford, Brad Lane, Vic Ciesielski, Felix Lipkin}
\institute{University of Melbourne, Parkville, Vic. 3010, Australia,\\
\email{mashford@unimelb.edu.au},\\ WWW home page:
\texttt{https://findanexpert.unimelb.edu.au/profile/11242-john-mashford}
\and
Commonwealth Scientific and Industrial Research Organisation \\
Private Bag 10, Clayton South, Vic. 3169, Australia \\
\and
 RMIT University \\
Computer Science and Software Engineering \\
GPO Box 2476, Melbourne, Vic. 3000, Australia}

\maketitle              

\begin{abstract}
This paper describes the results of formally evaluating the MCV (Markov concurrent vision) image labeling algorithm which is a (semi-) hierarchical algorithm commencing with a partition made up of single pixel regions and merging regions or subsets of regions using a Markov random field (MRF) image model. It is an example of a general approach to computer vision called concurrent vision in which the operations of image segmentation and image classification are carried out concurrently. While many image labeling algorithms output a single partition, or segmentation, the MCV algorithm outputs a sequence of partitions and this more elaborate structure may provide information that is valuable for higher level vision systems. With certain types of MRF the component of the system for image evaluation can be implemented as a hardwired feed forward neural network.  While being applicable to images (i.e. 2D signals), the algorithm is equally applicable to 1D signals (e.g. speech) or 3D signals (e.g. video sequences) (though its performance in such domains remains to be tested). The algorithm is assessed using subjective and objective criteria with very good results.
\end{abstract}
\section{Introduction}

The MCV algorithm was outlined in \cite{14,17,15}. The more or less current form is described in \cite{16,7}. It is part of the class of algorithms called ``deep learning" since the MCV partition trees generally have of the order of 10 layers. 

MRFs have been used for many years in computer vision applications \cite{9,29}. Panjwani and Healy \cite{37} describe an unsupervised texture segmentation algorithm based on agglomerative hierarchical clustering using MRFs. The clustering involves stepwise merging optimizing the conditional pseudolikelihood of the image. Wilson and Li \cite{35} describe a coarse to fine approach based on a quadtree process involving MRFs. Kato and Pong \cite{36} describe an MRF segmentation model based on Bayesian estimation using combinatorial optimization in the form of simulated annealing. 

Sharon {\em et al.} \cite{22} describe an approach that
utilizes hierarchical aggregation from a partition formed from single-pixel regions as in \cite{14}.

The main goal of the work described in the present paper was to evaluate the MCV algorithm on the basis of both subjective and objective criteria. The objective criteria considered are the principal criteria used by researchers for the quantitative  evaluation of region-based segmentation algorithms, that is, the probabilistic Rand index (PRI) and the variation of information (VOI). The present paper seeks to formally describe the motivation for MCV algorithm and the result of its evaluation.

The second section of this paper describes a general approach to hierarchical low-level vision based on representing the process of connected component labeling in terms of trees of partitions and then outlines some aspects of the MCV algorithm. Section 3 provides a formal evaluation of the MCV algorithm on the basis of both subjective (qualitative) and objective (quantitative) criteria. and the paper concludes with the final section.

\section{Hierarchical representation of pixel-based low level vision systems}

The simplest low-level vision system is formed by thresholding intensity and then carrying
out connected component labeling. More generally, a low level vision system may operate by
means of a pixel classifier $\gamma : X \rightarrow C$ where $X = \{1, \ldots, m\} \times \{1, \ldots, n\}$ is the image lattice and $C
\neq \emptyset$ is a finite set of classification labels e.g. for a simple foreground/background classification
$C = \{0,1\}$. The set of all regions of interest
(ROIs) or objects in the image is taken to be
\begin{equation}
\Pi = \cup\{\mbox{Comp}(\gamma^{-1}(c)) : c\in C\},
\end{equation}
where, for any $S \subset X$, Comp$(S)$ denotes the set of connected components of $S$. Thus the objects
in the image are taken to be the connected components of the sets of constant classification by
$\gamma$. The classifier $\gamma$ may act only on the pixel values of its argument or, more generally, it may
act on features extracted from a neighborhood of its argument. Since $\Pi_c = \mbox{Comp}(\gamma^{-1}(c))$ is a
partition of $\gamma^{-1}(c)$ for all $c \in C$ and $\gamma^{-1}(C) = \cup\{\gamma^{-1}(c) : c \in C\} = X$ it follows that $\Pi$ is a partition of $X$.

We may describe a hierarchical algorithm for generating $\Pi$ by describing a hierarchical algorithm
for generation of Comp$(S)$ for any $S \subset X$ and then applying the algorithm to $\gamma^{-1}(c)$ for all $c\in C$. For any set $\Omega$ let $P(\Omega)$ denote the power set of $\Omega$, thus $P(\Omega)$ (also denoted as $2^{\Omega}$) is the collection of all subsets of $\Omega$. Also let $P^2(\Omega) = P(P(\Omega))$. For $S \subset X$ define $S_0(x) = S \cap (x + W_0)$ where $W_0 \subset {\bf Z}^2$ (${\bf Z}$ denotes the integers) is the neighborhood of $0$ defining the neighbourhood relation
with respect to which connected components are being computed. For the work of this paper
we take $W_0$ to be the usual $8$-neighbourhood of the origin 0 (together with the point $0$).

For $S \subset X$ define the operator $M : S \times P^2(S) \rightarrow P^2(S)$ by
\begin{equation}
M(x,\Lambda) = \{{M_1(x,\Lambda)}\} \cup M_2(x,\Lambda),
\end{equation}
where
\begin{equation}
M_1(x,\Lambda) = \cup\{\lambda\in\Lambda : \lambda \cap S_0(x) \neq\emptyset\},
\end{equation}
\begin{equation}
M_2(x,\Lambda) = \{\lambda\in\Lambda : \lambda\cap S_0(x) = \emptyset\}.
\end{equation}

It is straightforward to show that the operator $M$ has the following properties: \newline
P1: If $x \in S$ and $\Lambda$ is a partition of $S$ then $M(x,\Lambda)$ is a partition of $S$ which is coarser than $\Lambda$. \newline
P2: If $x \in S$ and $\Lambda$ is a collection of connected regions then $M(x,\Lambda)$ is a collection of connected
regions.

We define the hierarchical algorithm for connected component labeling of a set $S\subset X$ as follows. \newline
Algorithm 1: \newline
Initialize $\Lambda= \{\{x\} : x \in S\}, S^{\prime} = S$ and then carry out the following procedure. \newline
While $S^{\prime}\neq\emptyset$,
\begin{enumerate}
\item Choose $x\in S^{\prime}$
\item Set $\Lambda := M(x,\Lambda), S^{\prime}:=S^{\prime}\backslash\{x\}$
\end{enumerate}

The algorithm must terminate because $|S| \leq|X|< \infty$. By Properties P1 and P2 the evolving
partition $\Lambda$ describes a tree of partitions of $S$ into connected sets.

We will show, using proof by contradiction, that after the termination of the algorithm $\Lambda = \mbox{Comp}(S)$. To this effect suppose that $\xi\in \Lambda_{\mbox{final}}$ and $\xi$ is not a connected component. Then $\exists\eta\supset\xi$
such that $\eta\neq\xi$ and $\eta\in\mbox{Comp}(S)$. Choose $x \in \xi, y \in \eta\backslash\xi$  such that $y \in S_0(x)$. Let $\Lambda$ be the evolving partition at the time when $x$ is chosen in Step 1 of Algorithm 1. Then $S_0(x) \subset M_1(x,\Lambda)$
and so $S_0(x)$ is contained in $\zeta$ for some $\zeta\in\Lambda_{\mbox{final}}$. Since $x \in \xi$ it follows that $\zeta=\xi$ and so $S_0(x) \subset\xi$ which contradicts the fact that $y {\slas \in}\xi$.

It follows from this result that the final partition of $S$ generated by Algorithm 1 is independent
of the choices that are made in executing the algorithm.

The MCV algorithm is formally described in \cite{16}.

One of the principal components of the MCV algorithm is the procedure for evaluating a region $R\subset W$, where $W\subset X$ is some window, for homogeneity resulting in, as the algorithm is currently implemented, in a simple YES or NO answer e.g. an element of $\{0,1\}$. In the algorithm in its current form the image on the square window $W$ containing $R$ is stepped down in resolution to the window $W_0$, being the fundamental neighbourhood of the origin in ${\bf Z}^2$. 

The step down procedure can be implemented by a hardwired pyramidal neural network. The image, $\omega$, say, in $W_0$ is then evaluated with respect to an autoregressive stochastic process defined in terms of a Gaussian Markov random field. The equilibrium distribution $\pi$ for the process can be used to evaluate the image on $W_0$. If $\pi(\omega)$ is large then the image $\omega$ is likely to result from the Gaussian MRF while if $\pi(\omega)$ is small then it is unlikely that $\omega$ would result from the MRF. The evaluation decision can be effected by setting a threshold on the probability which can equivalently be represented as a threshold on the energy function associated with the Gaussian MRF.  

The Gaussian Markov evaluation can be effected by a 2 layer perceptron hardwired neural network. Therefore the whole image evaluation process can be effected by a hardwired neural network. 

The output of the MCV system is a multiresolution sequence of partitions. After being processed
to determine additional structures such as feature vectors and classifications for the regions
in the partition sequence it may be passed to high level vision systems.

The high level vision system may select a level at which to process the output of MCV, treating it as a simple segmentation, or else it may carry out more elaborate processing on the multiresolution partition tree output by MCV.  

\section{Formal evaluation of the MCV algorithm}

The algorithm was tested on some images freely available on the internet, these being the images from the Berkeley segmentation dataset \cite{13}.

\subsection{Subjective criteria}

For the experiments described below the window $\Psi_i$ was taken to be a square of side length
$2r_i + 1$ where $r_i = 2^i$ pixels for $i = 1, \ldots, \mbox{max\_level}$. A random permutation of the pixels in the
image lattice can be computed offline and read from a file by the algorithm if it is to operate
on images of known fixed size, otherwise raster scan can be used. The algorithm was found to
be more efficient when a random permutation was used. Typically a good segmentation is
achieved using max\_levels = 9 with a random permutation as opposed to requiring
max\_levels = 11  if raster scan is used. This level of depth places the MCV algorithm within the general area of deep learning.

Some results of segmenting Google street view and Google earth images are given in \cite{16} and it can be seen that the resulting segmentations are excellent.

\begin{figure}
\centering
\includegraphics[width=0.75\textwidth]{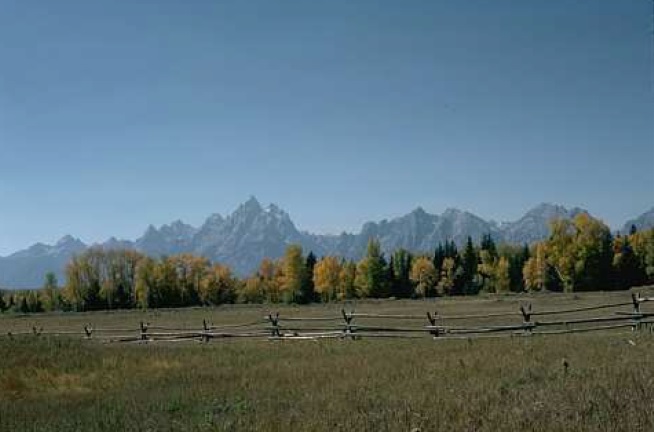}
\caption{Image from Berkeley segmentation database} \label{fig:Fig5}
\end{figure}

\begin{figure}
\centering
\includegraphics[width=0.75\textwidth]{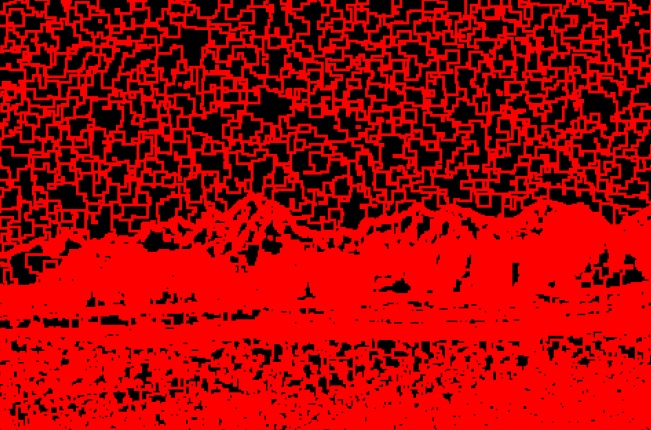}
\caption{Superpixels for Fig.~\ref{fig:Fig5} at level = 2} \label{fig:Fig6} 
\end{figure}

\begin{figure}
\centering
\includegraphics[width=0.75\textwidth]{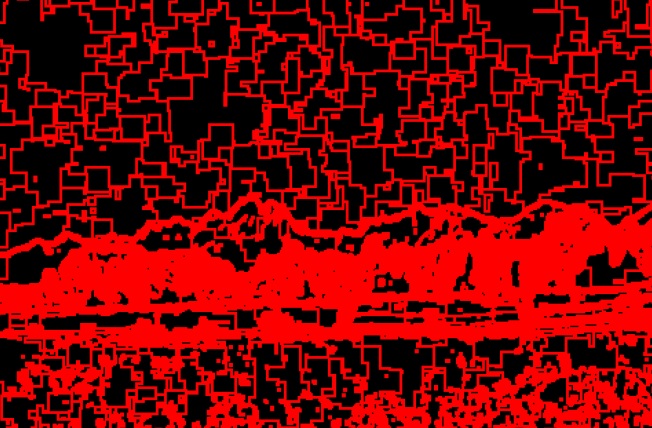}
\caption{Superpixels for Fig.~\ref{fig:Fig5} at level = 3} \label{fig:Fig7}
\end{figure}

\begin{figure}
\centering
\includegraphics[width=0.75\textwidth]{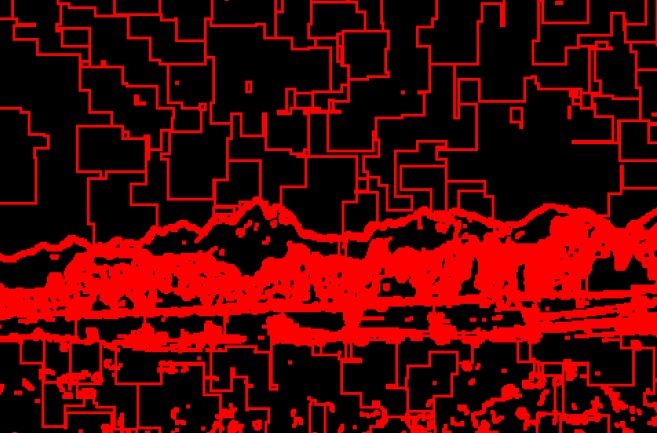}
\caption{Superpixels for Fig.~\ref{fig:Fig5} at level = 4} \label{fig:Fig8}
\end{figure}

\begin{figure}
\centering
\includegraphics[width=0.75\textwidth]{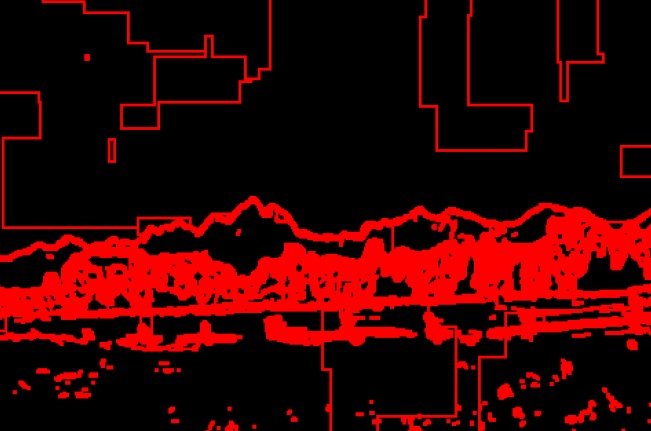}
\caption{Superpixels for Fig.~\ref{fig:Fig5} at level = 5} \label{fig:Fig9}
\end{figure}

\begin{figure}
\centering
\includegraphics[width=0.75\textwidth]{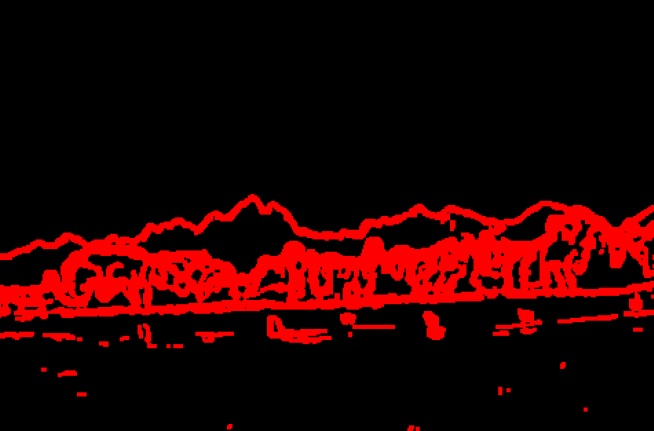}
\caption{Superpixels for Fig.~\ref{fig:Fig5} at level = 9} \label{fig:Fig10}
\end{figure}

\begin{figure}
\centering
\includegraphics[width=0.75\textwidth]{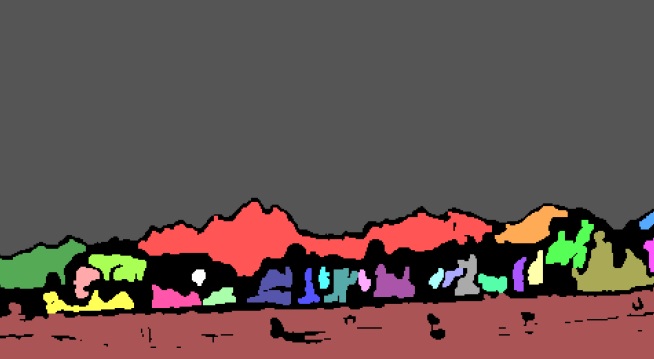}
\caption{Partition at level 10, final segmentation of image of Fig~\ref{fig:Fig5}} \label{fig:Fig11}
\end{figure}

The MCV algorithm results in a sequence of partitions each of which can be considered as a
collection of superpixels \cite{1}. A superpixel sequence corresponding to the image of Fig.~\ref{fig:Fig5} is given by the Figures~\ref{fig:Fig6} to~\ref{fig:Fig10} with the final MCV segmentation of the image being given by the image of Fig.~\ref{fig:Fig11}. 

It has been found experimentally that the results of the segmentation
are better when $\pi_{\mbox{pixels}}$ is random. This means that, in principle, the MCV algorithm is stochastic.
Different segmentations result when different random permutation of pixels are used. It
has been found experimentally that the differences are quite small, the Rand index \cite{24} of
pairs of segmentations obtained using different random pixel permutations is small. Further
work may involve formal investigation of this property from a theoretical point of view.

In summary, from these visual demonstrations, MCV performs excellently from a subjective point of view.

\subsection{Objective (quantitative) criteria}

The MCV algorithm was compared with state of the art region-based segmentation algorithms on the basis of the two most important criteria for comparing region-based segmentation algorithms, these being the probabilistic Rand index (PRI) \cite{24} and the variation of information (VOI) \cite{19}. While these criteria are quantitative they also have an element of subjectivity because the PRI and VOI scores are computed for the output segmentation with respect to segmentations produced by humans. 

It is important to note that there are many effective edge-based segmentation algorithms such as \cite{32} and these algorithms are assessed and compared using different criteria to those used for assessing region-based algorithms.

The comparison was made using the 200 images in the training set of the Berkeley segmentation dataset (BSD) \cite{13} and the 200 images of the test set of BSD.

Only one parameter is required to be specified in order for MCV to compute a partition sequence, that is the Markov threshold of the Gaussian MRF which evaluates images on the lowest resolution window $W_0$. If it is desired for MCV to produce a single output partition, i.e. a final segmentation, e.g. for the purpose of evaluating MCV with respect to evaluation criteria such as PRI and VOI, then it is necessary to also specify a level at which the segmentation is to be taken.  

\begin{figure}
\centering
\includegraphics[width=0.75\textwidth]{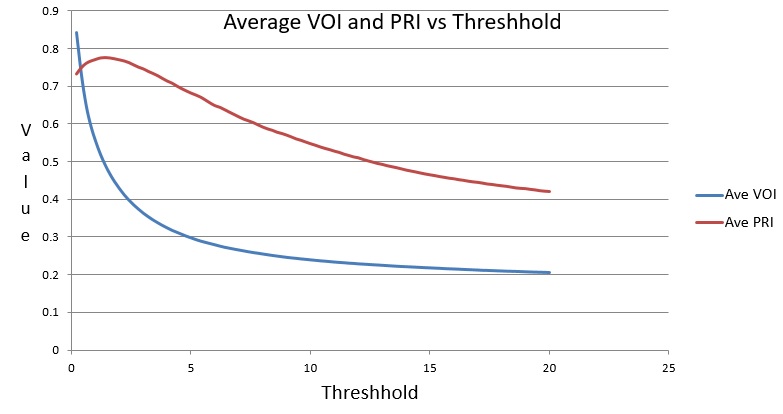}
\caption{Average PRI and average VOI versus threshold over BSD dataset} \label{fig:PRI_VOI}
\end{figure}

It was found that running the MCV algorithm on the training set resulted in excellent average PRI performance over BSD for many values of the (level, threshold) pair (see Table~\ref{tab:performance_table1}. Similarly, it was found that excellent average VOI performance over the BSD training set could be obtained for many values of the (level, threshold) pair (see Table~\ref{tab:performance_table2}). However it was found that the (level, threshold) pairs that did well with respect to PRI did not generally do well with respect to VOI.  This can be seen by examining Fig.~\ref{fig:PRI_VOI} which shows the variation of the average PRI and average VOI with threshold. PRI is favoured by low thresholds because it has higher values for lower thresholds while VOI is favoured by high thresholds because it has lower values for higher thresholds.

Nevertheless the overall average of max of PRI and the average of min of VOI over the BSD test set compare very well with the performance of many, if not all, of the state of the art segmentation algorithms as can be seen by examination of Table~\ref{tab:performance_table3} together with the method type legend table Table~\ref{tab:Method_type_legend_for_table_3}  (which may be compared with Table 2 of \cite{33}).

\section{Conclusion}

This paper has described the quantitive evaluation of the MCV image labeling algorithm (i.e. tables of numbers, to see the excellent performance of the algorithm visually please refer to \cite{16,7}). The algorithm is a (semi-) hierarchical
algorithm that may generate a simple segmentation partition or a multiresolution sequence
of partitions which can be useful to higher level vision systems. It utilizes an MRF image model in order to evaluate sub-images for homogeneity.
When certain MRF image models are used, such as a Gaussian MRF, the image evaluation procedure can be effected by a hardwired feed-forward neural
network. It is to be emphasized that this is a hardwired unsupervised network, the weights arise directly from the fixed MRF image model and training is not required. The algorithm executes very rapidly on the Berkely segmentation benchmark dataset
images (a few seconds per image) and the merge operation of the algorithm is massively parallelizable. Therefore, if parallelized and implemented using GPU, execution time should not be an issue. The algorithm generalizes to nD signals such as 1D (e.g. speech) or 3D (e.g. video) though its practical usefulness in such domains remains to be established. The MCV algorithm as we have described is fully unsupervised apart from using training to optimize the selection of (level, threshold) pairs. However it is expected that more effective scene understanding systems will form if the algorithm is generalized to a supervised algorithm in future work. Such a generalization may involve more sophisticated merge acceptance functions involving updating of merge confidence levels using Bayesian methodology.

\begin{table}[h]
\caption{Threshold, level and probabilistic Rand index (PRI) for MCV on the Berkeley segmentation training set ordered by PRI for the top 20 (threshold, level) pairs}\label{tab:performance_table1}
\begin{center}
\begin{tabular}{lll}
\hline
Threshold	&Level		&Average PRI  \\
       \hline
2.4		&12		&0.787853 \\
2.2		&12		&0.787536 \\
2.8		&9		&0.787353 \\
2.4		&9		&0.787265 \\
2		&15		&0.786808 \\
3		&9		&0.786751 \\
2.6		&9		&0.786276 \\
1.4		&21		&0.786157 \\
2		&12		&0.786087 \\
1.8		&12		&0.785809 \\
1.6		&18		&0.785684 \\
1.2		&18		&0.785352 \\
1.6		&15		&0.785231 \\
2.6		&12		&0.785201 \\
1		&24		&0.785194 \\
1.8		&18		&0.785064 \\
1.8		&15		&0.785018 \\
1		&33		&0.784838 \\
1		&30		&0.784838 \\
3.2		&9		&0.7848 \\
\hline 
\end{tabular}
\end{center}
\end{table}

\begin{table}[h]
\caption{Threshold, level and variation of information (VOI) for MCV on the Berkeley segmentation training set ordered by VOI for the top 20 (threshold, level) pairs}\label{tab:performance_table2}
\begin{center}
\begin{tabular}{lll}
\hline
Threshold	&Level		&Average VOI  \\
\hline
20		&33		&1.80146 \\
20		&30		&1.80146 \\
19.8		&33		&1.8047 \\
19.8		&30		&1.8047 \\
19.6		&33		&1.80771 \\
19.6		&30		&1.80771 \\
19.4		&33		&1.81073 \\
19.4		&30		&1.81073 \\
19.2		&33		&1.81395 \\
19.2		&30		&1.81395 \\
19		&33		&1.81736 \\
19		&30		&1.81736 \\
20		&27		&1.81752 \\
18.8		&33		&1.82058 \\
18.8		&30		&1.82058 \\
19.8		&27		&1.8206 \\
19.6		&27		&1.82389 \\
18.6		&33		&1.82424 \\
18.6		&30		&1.82424 \\
19.4		&27		&1.82751 \\
\hline
\end{tabular}
\end{center}
\end{table}

\begin{table}[h]
\caption{Probabilistic Rand index (PRI) and variation of information (VOI) for MCV compared to various state of the art segmentation algorithms. The rank order of each algorithm with respect to each criterion is shown in brackets.}\label{tab:performance_table3}
\begin{center}
\begin{tabular}{cll}
\hline
       Method &PRI &VOI \\
       \hline
        MCV         &0.829285 (ave max)(2)    &1.78093 (ave min)(3)  \\
        Ncut         &0.7242(11)    &2.9061(10)   \\
        MNcut      &0.7559(9)    &2.4701(7)  \\
        SAS          &0.8319(1)    &1.6849(1) \\
        FusionTP  &0.7771(4)    &3.3089(11)  \\
        NTP          &0.7521(10)    &2.4954(8) \\
        KmsGC     &0.7712(7)    &2.5616(9)  \\
        JSEG        &0.7756(6)    &2.3217(5) \\
        CTM         &0.7561(8)    &2.4640(6) \\
        TBES        &0.8070(3)    &1.7050(2) \\
        Yin {\em et al.}     &0.7769(5)    &2.3067(4) \\
\hline 
\end{tabular}
\end{center}
\end{table}

\begin{table}[h]
\caption{Method type legend for Table~\ref{tab:performance_table3}\label{tab:Method_type_legend_for_table_3}}
\begin{center}
\begin{tabular}{cc}
\hline
       Method &Reference \\
       \hline
        Ncut          &Shi and Malik \cite{23}   \\
        MNcut       &Cour and Benezit \cite{6}  \\
        SAS          &Li, Wu and Chang \cite{31} \\
        FusionTP  &Zhou, Bai {\em et al.} \cite{30}  \\
        NTP          &Wang, Jia {\em et al.} \cite{25} \\
        KmsGC    &Liang, Zhang {\em et al.} \cite{11}  \\
        JSEG          &Wang, Tang {\em et al.} \cite{26} \\
        CTM          &Yang, Wright {\em et al.} \cite{27} \\
        TBES          &Mobahi, Rao {\em et al.}  \cite{20} \\
        Yin {\em et al.}   &Yin {\em et al.}  \cite{33} \\
\hline 
\end{tabular}
\end{center}
\end{table}

\section*{Acknowledgments}

The work described in this paper was partially funded by the Commonwealth Scientific and Industrial Research Organisation (CSIRO, Australia). Also the authors would like to thank Mike Rahilly, Lachlan McAlpine and Geoff Bryan for help with this work.

\end{document}